\documentclass[preprint,3p,11pt]{elsarticle}

\usepackage[utf8]{inputenc}
\usepackage[T1]{fontenc}
\usepackage{amsmath,amssymb}
\usepackage{graphicx}
\usepackage{booktabs}
\biboptions{numbers,sort&compress}
\usepackage{hyperref}
\usepackage{lineno}
\usepackage{xcolor}
\usepackage{multirow}
\usepackage{caption}
\usepackage{siunitx}

\journal{Remote Sensing of Environment}

\begin{document}

\begin{frontmatter}

\title{Depth-Resolved Coral Reef Thermal Fields from Satellite SST and Sparse In-Situ Loggers Using Physics-Informed Neural Networks}

\author[cse,caid]{Alzayat Saleh\corref{cor1}}
\ead{alzayat.saleh@my.jcu.edu.au}
\author[cse,caid]{Mostafa Rahimi Azghadi}

\cortext[cor1]{Corresponding author.}

\affiliation[cse]{organization={College of Science and Engineering,
  James Cook University},
  city={Townsville, QLD},
  country={Australia}}
\affiliation[caid]{organization={Centre for AI and Data Science Innovation,
  James Cook University},
  city={Townsville},
  country={Australia}}

\begin{abstract}
Satellite sea surface temperature (SST) products underpin global coral bleaching monitoring, yet they measure only the ocean skin.
Corals inhabit depths from the shallows to beyond \SI{20}{\metre}, where temperatures can be 1--\SI{3}{\degreeCelsius} cooler than the surface; applying satellite SST uniformly to all depths therefore overestimates subsurface thermal stress.
We present a physics-informed neural network (PINN) that fuses NOAA Coral Reef Watch SST with sparse in-situ temperature loggers within the one-dimensional vertical heat equation, enforcing SST as a hard surface boundary condition and jointly learning effective thermal diffusivity ($\kappa$) and light attenuation ($K_d$).
Validated across four Great Barrier Reef sites (30 holdout experiments), the PINN achieves 0.25--\SI{1.38}{\degreeCelsius} RMSE at unseen depths.
Under extreme sparsity (three training depths), the PINN maintains \SI{0.27}{\degreeCelsius} RMSE at the \SI{5}{\metre} holdout and \SI{0.32}{\degreeCelsius} at the \SI{9.1}{\metre} holdout, where statistical baselines collapse to $>$\SI{1.8}{\degreeCelsius}; it outperforms a physics-only finite-difference baseline in 90\% of experiments.
Depth-resolved Degree Heating Day (DHD) profiles show that thermal stress attenuates with depth: at Davies Reef, DHD drops from 0.29 at the surface to zero by \SI{10.7}{\metre}, consistent with logger observations, while satellite DHD remains constant at 0.31 across all depths.
However, the PINN underestimates absolute DHD at shallow depths because its smooth predictions attenuate the short-duration peaks that drive threshold exceedances; PINN DHD values should be interpreted as conservative lower bounds on depth-resolved stress.
These results demonstrate that physics-constrained fusion of satellite SST with sparse loggers can extend bleaching assessment to the depth dimension using existing observational infrastructure.
\end{abstract}

\begin{keyword}
physics-informed neural network \sep
coral bleaching \sep
satellite SST \sep
depth-resolved temperature \sep
Great Barrier Reef \sep
thermal stress \sep
degree heating days \sep
data fusion
\end{keyword}


\end{frontmatter}

\section{Introduction}
\label{sec:introduction}

Mass coral bleaching has shifted from a rare disturbance to a recurring crisis.
Since 2016, the Great Barrier Reef (GBR) alone has experienced six mass bleaching events (2016, 2017, 2020, 2022, 2024, and 2025), each triggered by marine heatwaves that push reef temperatures above coral thermal tolerance thresholds for sustained periods \citep{hughes2017global, hughes2018spatial, gbrmpa2025bleaching}.
Satellite sea surface temperature (SST) products, principally the NOAA Coral Reef Watch (CRW) system, are the backbone of global bleaching monitoring: CRW delivers daily 5\,km SST and cumulative thermal stress metrics (Degree Heating Weeks) for every reef area on Earth \citep{liu2014reef, skirving2019relentless}.
These products have transformed our ability to detect and forecast bleaching events at planetary scale.

Satellite SST, however, measures only the skin temperature of the ocean surface.
Corals inhabit a range of depths, from reef flats at less than \SI{1}{\metre} to reef slopes and walls extending beyond \SI{20}{\metre}.
In-situ measurements consistently reveal that temperature decreases with depth on coral reefs, with subsurface temperatures \SIrange{1}{3}{\degreeCelsius} cooler than the surface during thermal stress events \citep{leichter2006variation, baird2018decline}.
This vertical thermal gradient means that corals at \SIrange{10}{20}{\metre} may experience far less thermal stress than satellite observations suggest.
Bleaching alerts derived from satellite SST alone therefore overestimate thermal stress at depth, a bias that affects management decisions about depth refugia, reef prioritization, and the interpretation of bleaching surveys \citep{bridge2013depth, frade2018deep, bongaerts2010assessing}.

\begin{figure}[!htbp]
  \centering
  \includegraphics[width=\textwidth]{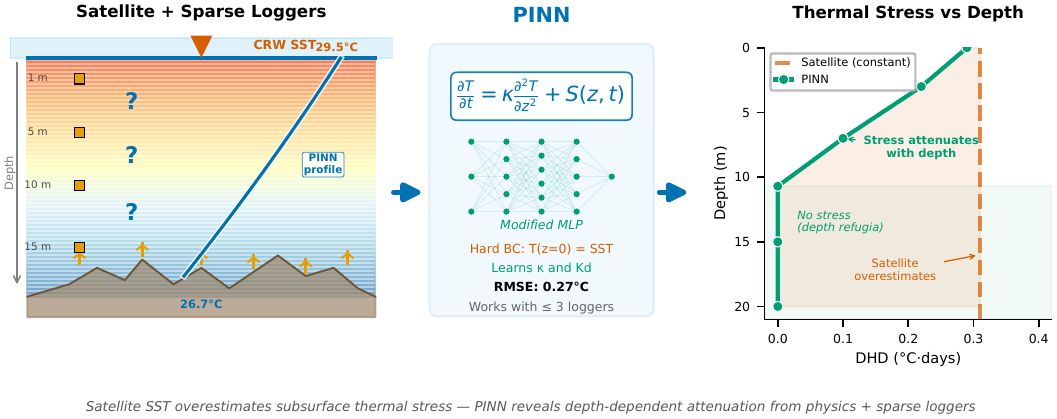}
  \caption{Conceptual overview of the PINN framework. \textbf{Left}: Satellite SST provides a surface boundary condition while sparse AIMS temperature loggers supply subsurface observations at discrete depths, leaving gaps in between. \textbf{Center}: A physics-informed neural network embeds the 1D vertical heat equation as a soft constraint, enforces the satellite SST as a hard boundary condition at the surface, and jointly learns effective thermal diffusivity ($\kappa$) and light attenuation ($K_d$). \textbf{Right}: The trained PINN produces continuous depth-resolved temperature and Degree Heating Day (DHD) profiles, revealing that thermal stress attenuates with depth; this vertical structure is invisible to satellite-only monitoring.}
  \label{fig:conceptual_overview}
\end{figure}

Resolving this depth bias requires depth-resolved thermal fields, but constructing them from observations alone is difficult.
In-situ temperature loggers provide point measurements at specific depths, yet even well-instrumented networks such as the AIMS temperature logger program on the GBR cover only tens of sites with varying depth coverage \citep{bainbridge2017temperature}.
Reconstructing continuous thermal fields across the full depth range from these sparse observations demands a method that can honor the measurements, enforce physical consistency between depths, and use satellite SST as a reliable surface anchor.
Standard interpolation methods (inverse distance weighting, nearest neighbor, Gaussian process regression) can fill gaps within the observed depth range but lack physical constraints; they degrade rapidly when asked to extrapolate beyond training data or when observations become sparse.

Physics-informed neural networks (PINNs) offer a framework for this problem.
PINNs embed governing physical equations directly into the neural network training process, producing solutions that simultaneously fit observed data and approximately satisfy the underlying physics \citep{raissi2019physics, karniadakis2021physics}.
For depth-resolved temperature reconstruction, the relevant physics is the vertical heat equation: temperature evolves through a balance of vertical diffusive mixing and depth-attenuated solar heating.
By encoding this equation as a soft constraint, a PINN can interpolate and extrapolate temperature fields that are physically plausible even where no observations exist.
This is particularly valuable for coral reef applications, where the physics is well understood but observations are sparse.
PINNs have demonstrated success in related geophysical inverse problems, including subsurface flow reconstruction and ocean state estimation \citep{kashinath2021physics, willard2022integrating}.
Recently, PINNs have been applied to subsurface ocean temperature reconstruction from surface observations \citep{xiao2026observation, han2025application}, though none have targeted coral reef environments or the specific challenge of depth-resolved thermal stress assessment.
Figure~\ref{fig:conceptual_overview} illustrates the approach we develop here: satellite SST and sparse in-situ loggers are fused through a PINN constrained by the vertical heat equation, producing continuous depth-resolved temperature and thermal stress profiles.

This paper makes three contributions:
\begin{enumerate}
  \item A PINN framework for reconstructing depth-resolved thermal fields from satellite SST and sparse temperature loggers, enforcing the one-dimensional heat equation with depth-dependent solar heating and jointly learning effective thermal diffusivity ($\kappa$) and light attenuation ($K_d$) from data.
  \item Holdout validation across four GBR sites, demonstrating reconstruction accuracy ranging from 0.25 to \SI{1.38}{\degreeCelsius} RMSE and stability under extreme data sparsity where statistical baselines collapse.
  \item Depth-resolved Degree Heating Day profiles that reveal vertical attenuation of thermal stress, with matched-window logger validation demonstrating that the PINN captures the qualitative depth structure of DHD while providing conservative absolute estimates.
\end{enumerate}

The remainder of this paper is organized as follows.
Section~\ref{sec:study_area} describes the four GBR study sites and the satellite and in-situ data sources.
Section~\ref{sec:methods} details the PINN formulation, network architecture, baseline methods, and experimental design.
Section~\ref{sec:results} presents holdout validation, sparsity robustness, depth-resolved thermal stress comparisons, and learned physical parameters.
Section~\ref{sec:discussion} discusses implications for bleaching monitoring, the PINN's advantages and limitations, and future directions.
Section~\ref{sec:conclusions} summarizes the findings.

\section{Study area and data}
\label{sec:study_area}

\subsection{Great Barrier Reef study sites}
\label{sec:sites}

\begin{figure}[!htbp]
  \centering
  \includegraphics[width=0.9\columnwidth]{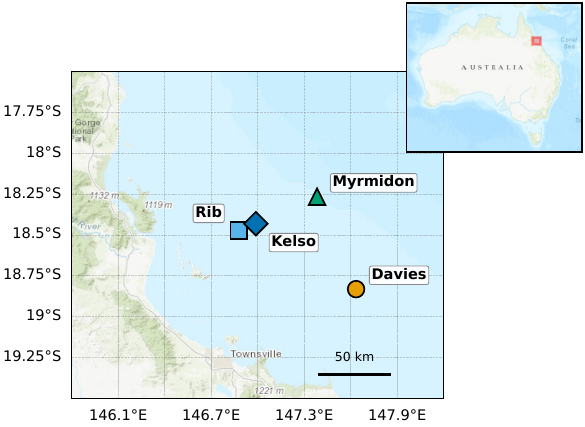}
  \caption{Study area showing the four Great Barrier Reef sites in the central GBR.
    Inset shows location on the northeast Australian coast. Each site is instrumented
    with AIMS temperature loggers at multiple depths. Archive-wide depth counts
    (Davies: 30, Myrmidon: 19, Rib: 14, Kelso: 12) differ from the depths
    available in each study window (Table~\ref{tab:data_summary}). Map lines delineate study areas and do not necessarily depict accepted national boundaries. Basemap: Esri World Topographic Map.}
  \label{fig:study_area}
\end{figure}

We selected four reefs in the central GBR that span a range of shelf positions, depth ranges, and bleaching histories (Fig.~\ref{fig:study_area}; Table~\ref{tab:data_summary}).
Davies Reef (\SI{18.83}{\degree}S, \SI{147.63}{\degree}E) is a mid-shelf platform reef with extensive instrumentation across 30 depth levels in the full archive, providing the densest vertical coverage in our study.
Myrmidon Reef (\SI{18.27}{\degree}S, \SI{147.38}{\degree}E) is an outer-shelf reef with a deeper water column and lower wave exposure; its study window (2020--2024) encompasses the 2020 and 2024 mass bleaching events.
Rib Reef (\SI{18.47}{\degree}S, \SI{146.88}{\degree}E) is a mid-shelf reef where loggers recorded active bleaching stress during the 2020 and 2022 events, with temperature coverage to \SI{9}{\metre}.
Kelso Reef (\SI{18.43}{\degree}S, \SI{146.99}{\degree}E) provides a historical baseline: its 1998--2001 time window captures the major 1998 mass bleaching event, with loggers to \SI{19}{\metre}.

\begin{table}[!htbp]
  \centering
  \caption{Summary of temperature logger data at each reef. ``Archive depths'' is the total across the full multi-decade AIMS archive; ``Window depths'' is the number of unique depths with data in the selected study period. Observations are after quality control and temporal aggregation. $^\dagger$Nearest mass bleaching event; occurs after the Davies study window.}
  \label{tab:data_summary}
  \begin{tabular}{lccccccc}
    \toprule
    Reef & Period & Duration & Archive & Window & Obs. & Max depth & Bleaching \\
         &        & (yr)     & depths  & depths & ($\times 10^6$) & (m) & events \\
    \midrule
    Davies   & 2011--2014 & 3.6 & 30 & 20 & $\sim$15 & $\sim$20 & 2016$^\dagger$ \\
    Myrmidon & 2020--2024 & 4.3 & 19 &  6 & $\sim$12 & $\sim$20 & 2020, 2024 \\
    Rib      & 2018--2023 & 5.3 & 14 &  3 & $\sim$8  & $\sim$9  & 2020, 2022 \\
    Kelso    & 1998--2001 & 3.3 & 12 &  3 & $\sim$5  & $\sim$19 & 1998 \\
    \bottomrule
  \end{tabular}
\end{table}

\subsection{AIMS temperature logger data}
\label{sec:aims_data}

Temperature observations come from the Australian Institute of Marine Science (AIMS) temperature logger network, one of the most comprehensive long-term reef monitoring programs globally \citep{bainbridge2017temperature}.
The full archive for our four reefs comprises 277 CSV files containing approximately 55 million quality-controlled observations at 5--30 minute sampling intervals.
Two instrument types contribute data: weather station sensors with 30-column records and standalone loggers with 19-column records.
Depth is extracted from metadata columns or from the filename convention (e.g., \texttt{@5m}).

For each reef we selected a contiguous time window with maximum depth coverage (Table~\ref{tab:data_summary}).
Within each window, observations were aggregated to hourly means for PINN training, retaining sub-daily resolution for Degree Heating Day computation.
Quality flags provided by AIMS were applied to exclude suspect readings.

\subsection{Satellite SST}
\label{sec:satellite_sst}

We use the NOAA Coral Reef Watch daily SST product at \SI{5}{\kilo\metre} resolution \citep{liu2014reef}, accessed via ERDDAP (dataset \texttt{NOAA\_DHW}, variable \texttt{CRW\_SST}).
CRW SST provides the surface boundary condition for the PINN: at $z = 0$, the predicted temperature is constrained to equal the satellite observation exactly.
CRW SST is also used to compute the Maximum of Monthly Means (MMM) climatology at each reef, from which bleaching thresholds (MMM + \SI{1}{\degreeCelsius}) are derived for Degree Heating Day calculations.
This dual role makes satellite SST central to both the reconstruction method and the thermal stress assessment.

\section{Methods}
\label{sec:methods}

\subsection{Governing physics: the vertical heat equation}
\label{sec:heat_equation}

The vertical temperature structure on a coral reef is governed by the balance between turbulent diffusive mixing and depth-attenuated solar heating.
We model this as a one-dimensional heat equation:
\begin{equation}
  \frac{\partial T}{\partial t}
  = \kappa \, \frac{\partial^2 T}{\partial z^2}
  + \frac{Q_{\text{solar}}(z,t)}{\rho \, c_p}
  \label{eq:heat}
\end{equation}
where $T(z,t)$ is temperature (\si{\degreeCelsius}) at depth $z$ (\si{\metre}) and time $t$ (\si{\second}), $\kappa$ is the effective thermal diffusivity (\si{\metre\squared\per\second}), $\rho \, c_p = \SI{4.1e6}{\joule\per\metre\cubed\per\kelvin}$ is the volumetric heat capacity of seawater, and $Q_{\text{solar}}$ is the volumetric solar heating rate (Eq.~\ref{eq:solar}), derived from Beer's Law as the depth derivative of the downwelling irradiance:
\begin{equation}
  Q_{\text{solar}}(z,t) = Q_{\max} \, K_d \exp(-K_d \, z) \; f_{\text{diurnal}}(t)
  \label{eq:solar}
\end{equation}
Here $Q_{\max} = \SI{350}{\watt\per\metre\squared}$ is the peak surface irradiance, $K_d$ (\si{\per\metre}) is the light attenuation coefficient, and $f_{\text{diurnal}}(t) = \max\!\big(0, \sin(\pi(h_{\text{local}} - 6)/12)\big)$ restricts heating to daytime hours, where $h_{\text{local}}$ is the local solar hour (UTC$+$10 for the GBR).

Three physical parameters are treated as learnable, estimated jointly with the network weights during training: the base diffusivity $\kappa_0$, a depth coefficient $\alpha$ such that $\kappa(z) = \kappa_0 \exp(\alpha z)$, and the light attenuation coefficient $K_d$.
The log-linear depth dependence allows the effective diffusivity to vary with depth, capturing the tendency for turbulent mixing to differ between the wave-influenced surface layer and the calmer deep water.
The effective diffusivity represents the combined effect of turbulent mixing and molecular diffusion; it is not a measurement of either individually but an effective parameter that the heat equation requires to reproduce the observed vertical temperature evolution.
On coral reefs, turbulent diffusivities typically range from $10^{-4}$ to $10^{-3}$~\si{\metre\squared\per\second} \citep{monismith2007hydrodynamics, lowe2015oceanic}.

This formulation deliberately neglects lateral advection, tidal pumping, and surface cooling terms (longwave radiation, evaporative heat loss).
For reef-scale vertical profiles where vertical mixing dominates the thermal structure, the one-dimensional simplification is appropriate: lateral temperature gradients across a single reef are small relative to vertical gradients, and the omitted surface cooling terms are absorbed into the effective $\kappa$.
We revisit these simplifications in Section~\ref{sec:limitations}.

\subsection{PINN architecture and training}
\label{sec:architecture}

\subsubsection{Hard boundary condition}
The PINN enforces satellite SST as an exact surface boundary condition through a structural constraint on the network output (Eq.~\ref{eq:hard_bc}):
\begin{equation}
  T(z,t) = \text{SST}(t) + \frac{z}{z_{\max}} \; T_{\text{scale}} \; f_{\text{nn}}(z,t)
  \label{eq:hard_bc}
\end{equation}
where $f_{\text{nn}}$ is the raw neural network output, $z_{\max}$ is the maximum depth, and $T_{\text{scale}}$ is set from the observed temperature range.
At $z = 0$, the second term vanishes and $T(0,t) = \text{SST}(t)$ exactly, regardless of the network weights.
This construction is analogous to a surveyor's benchmark: just as elevation measurements are anchored to a known datum, the depth-resolved temperature field is anchored to the satellite observation at the surface.
The neural network learns only the depth-dependent correction, which grows with $z$.

\subsubsection{Modified MLP with multiplicative gating}
We use a modified multilayer perceptron (MLP) with multiplicative gating \citep{wang2021understanding} rather than a standard feedforward architecture.
Two encoding projections $U = \sigma(x W_U + b_U)$ and $V = \sigma(x W_V + b_V)$ are computed once from the raw input $x$.
At each hidden layer, the transformed representation $H$ is combined with these encodings via $H \leftarrow (1 - H) \odot U + H \odot V$, where $\odot$ denotes element-wise multiplication.
This gating mechanism allows each layer to interpolate between two learned representations of the input, enabling the network to capture interactions between depth attenuation and temporal variability more effectively than a standard MLP.
In preliminary testing, the modified MLP architecture substantially reduced holdout RMSE at deep holdout depths compared to a standard MLP with the same layer count and width. 

The network takes $(z, t)$ as input, where time is encoded using 8 diurnal Fourier harmonics (keyed to local hour) and 3 annual harmonics (keyed to day of year), yielding 22 periodic features that capture both sub-daily and seasonal patterns.
Five hidden layers of 128 units with $\tanh$ activation produce the output $f_{\text{nn}}(z,t)$.

\subsubsection{Loss function}
The PINN is trained by minimizing a composite loss (Eq.~\ref{eq:loss}):
\begin{equation}
  \mathcal{L} = w_{\text{data}} \, \mathcal{L}_{\text{data}}
              + w_{\text{pde}}  \, \mathcal{L}_{\text{pde}}
              + w_{\text{bc}}   \, \mathcal{L}_{\text{bc}}
  \label{eq:loss}
\end{equation}
where $\mathcal{L}_{\text{data}}$ is the mean squared error between predictions and logger observations, $\mathcal{L}_{\text{pde}}$ is the non-dimensionalized mean squared residual of Eq.~\eqref{eq:heat} evaluated at depth-stratified collocation points (scaled by $z_{\max}^2 / \kappa_{\text{ref}}$ and weighted by $\exp(-z/20)$ to emphasize near-surface physics), and $\mathcal{L}_{\text{bc}}$ enforces a Neumann (zero-flux) boundary condition at the bottom of the domain ($\partial T / \partial z \approx 0$ near $z = z_{\max}$).
Collocation points are depth-stratified: 50\% are sampled uniformly across the domain, and 50\% are concentrated within $\pm\SI{1}{\metre}$ of logger depths to ensure adequate PDE sampling near data constraints.
The surface boundary condition is enforced exactly through Eq.~\eqref{eq:hard_bc} and does not appear in the loss.

The PDE weight follows a linearly decreasing schedule: $w_{\text{pde}}$ starts at 1.0 and decreases to 0.5, 0.2, and 0.05 at epochs 3000, 6000, and 10000 respectively.
This curriculum strategy reflects a fundamental asymmetry: without early physics enforcement, the network can converge to interpolation solutions that satisfy the data loss but violate the heat equation, producing non-physical temperature profiles between observed depths.
Once the network has learned a physically plausible vertical structure, reducing the PDE weight allows fine-grained fitting of observations that the simplified 1D equation cannot fully capture (e.g., advective intrusions, tidal mixing events).
The specific schedule values were chosen empirically based on convergence behavior across preliminary experiments.

Table~\ref{tab:hyperparams} summarizes all training hyperparameters.

\begin{table}[!htbp]
  \centering
  \caption{PINN training hyperparameters.}
  \label{tab:hyperparams}
  \begin{tabular}{lll}
    \toprule
    Parameter & Value & Rationale \\
    \midrule
    Hidden layers            & 5 $\times$ 128   & Sufficient capacity \\
    Activation               & $\tanh$           & Best in A/B testing \\
    Fourier features         & 8 diurnal + 3 annual & Temporal periodicity \\
    Epochs                   & 15\,000           & Convergence verified \\
    Learning rate            & $10^{-3}$         & Cosine schedule \\
    Optimizer                & Adam              & Standard for PINNs \\
    Data batch size          & 5\,000            & Random subsampling \\
    Collocation points       & 2\,500            & Depth-stratified \\
    Collocation resampling   & Every 2\,000 ep.  & Diversity \\
    $w_{\text{data}}$        & 10.0              & Data fidelity \\
    $w_{\text{pde}}$ schedule & $1.0 \to 0.05$   & Curriculum \\
    $w_{\text{bc}}$            & 0.1               & Bottom Neumann BC ($\partial T/\partial z = 0$) \\
    $T_{\text{scale}}$       & Adaptive           & $\max(1.0,\; 0.5 \times \Delta T_{\text{obs}})$ \\
    Gradient clipping        & 1.0 (global norm) & Stability \\
    \bottomrule
  \end{tabular}
\end{table}

The framework is implemented in JAX \citep{bradbury2018jax}, enabling GPU-accelerated automatic differentiation for both forward predictions and PDE residual computation.
Training takes approximately 3.5--5 minutes per reef on an NVIDIA RTX~4090.

\subsection{Baseline methods}
\label{sec:baselines}

We compare the PINN against four statistical baselines, a physics-only numerical baseline, and the satellite-only reference, all operating on the same $(z, t) \to T$ mapping with identical train/test splits:

\begin{itemize}
  \item \textbf{Gaussian Process (GP):} Scikit-learn implementation with radial basis function plus white noise kernel, trained on $(z, t)$ inputs (subsampled to 500 points due to the $O(n^3)$ computational cost of exact GP inference; sparse GP approximations could potentially improve this baseline).
  \item \textbf{Inverse Distance Weighting (IDW):} Power-2 weighting using the 20 nearest neighbors in normalized $(z, t)$ space.
  \item \textbf{Nearest Neighbor (NN):} 1-nearest-neighbor in normalized $(z, t)$ space.
  \item \textbf{Random Forest (RF):} 200 trees (max depth 20, minimum 5 samples per leaf) with time-encoded features ($z$, diurnal sin/cos, normalized~$t$).
  \item \textbf{Finite-Difference (FD):} Numerical solution of the same 1D heat equation (Eq.~\ref{eq:heat}) via implicit Euler on a $100 \times N_t$ grid (hourly timesteps), using real CRW SST as the surface Dirichlet boundary condition, an insulating (zero-flux) bottom boundary, and literature-value parameters ($\kappa = \SI{2.5e-4}{\metre\squared\per\second}$, $K_d = \SI{0.1}{\per\metre}$, $Q_{\max} = \SI{350}{\watt\per\metre\squared}$, $\rho c_p = \SI{4.1e6}{\joule\per\metre\cubed\per\kelvin}$). This baseline isolates whether the neural network adds value beyond the physics alone.
  \item \textbf{Satellite-only:} CRW SST applied uniformly to all depths, representing current operational practice.
\end{itemize}

The four statistical baselines incorporate no physical constraints, while the FD baseline uses the same physics as the PINN but with fixed parameters and no data assimilation beyond the surface boundary condition.
The comparison therefore isolates the separate contributions of physics and data-driven learning.

\subsection{Experimental design}
\label{sec:experimental_design}

\subsubsection{Holdout validation}
For each reef, we systematically hold out one depth at a time, train all methods on the remaining depths, and evaluate predictions at the held-out depth.
This protocol tests each method's ability to reconstruct temperature at a depth it has never observed, the central challenge for extending sparse monitoring networks.

\subsubsection{Sparsity experiments}
We progressively reduce the number of training depths to test how each method degrades as monitoring coverage decreases.
Davies Reef, with the densest depth coverage (20 window depths), is tested at 5 sparsity levels (17, 10, 5, 3, and 2 training depths) for each of its 3 holdout depths, yielding 15 experiments.
When reducing from the full depth set, remaining training depths are selected at approximately evenly spaced intervals across the available depths (excluding the holdout), maximizing vertical coverage.
Myrmidon (6 window depths) is tested at 3 sparsity levels (5, 3, 2) for each of 3 holdout depths (9 experiments).
Rib and Kelso, with only 3 window depths each, support a single sparsity level (2 training depths) per holdout, yielding 3 experiments each.
The total experimental design comprises 30 holdout configurations.
Depths within \SI{0.3}{\metre} of the holdout are excluded from training to prevent data leakage, so the maximum effective training count at Davies is 17 (not 20).

\subsubsection{Degree Heating Days}
We compute Degree Heating Days (DHD) as the cumulative thermal stress above the bleaching threshold (Eq.~\ref{eq:dhd}):
\begin{equation}
  \text{DHD} = \sum_{t} \max\!\big(0,\; T(z,t) - T_{\text{threshold}}\big) \, \Delta t
  \label{eq:dhd}
\end{equation}
where $T_{\text{threshold}} = \text{MMM} + \SI{1}{\degreeCelsius}$ and MMM is the Maximum of Monthly Means computed from CRW SST at each reef \citep{liu2006overview}.
DHD is computed in units of \si{\degreeCelsius\,days}.

To ensure fair comparison, we compute logger, PINN, and satellite DHD over identical time windows at each depth.
The window at each depth corresponds to the deployment period of the logger at that depth.
This matched-window approach eliminates temporal sampling biases that would otherwise confound DHD comparisons across depths with different deployment durations.
Because logger observations are recorded at 5--30 minute intervals while PINN predictions are hourly, logger DHD captures sub-hourly thermal spikes that the PINN cannot resolve.
This temporal resolution mismatch contributes to differences between logger-observed and PINN-reconstructed DHD values, independent of prediction accuracy.

\section{Results}
\label{sec:results}

\subsection{Holdout validation across four reefs}
\label{sec:holdout_results}

\begin{figure*}[!htbp]
  \centering
  \includegraphics[width=\textwidth]{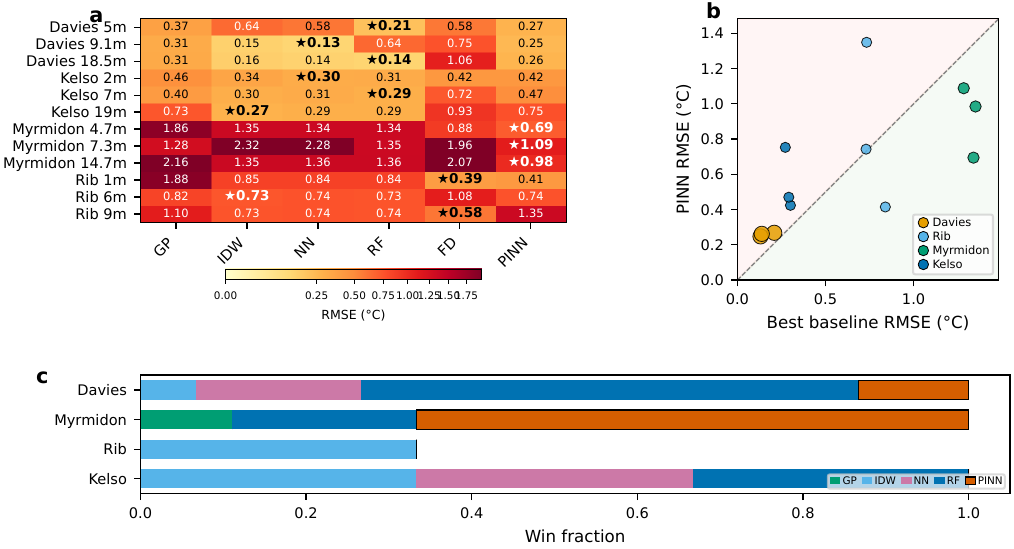}
  \caption{Holdout validation results across four reefs. (a)~Holdout RMSE
    (\si{\degreeCelsius}) for each method across all reef--depth combinations
    at full training depth; FD denotes the physics-only finite-difference baseline.
    Stars mark the best method per row.
    (b)~PINN RMSE vs.\ best baseline RMSE for each holdout experiment. Points
    below the diagonal indicate PINN superiority. Point size scales with the
    number of training depths; colors indicate reef identity.
    (c)~Win fraction by reef (horizontal stacked bars): proportion of holdout
    experiments won by each method. PINN RMSE values are means across 5 random seeds; standard
    deviations are typically below \SI{0.01}{\degreeCelsius}
    (median $\pm$\SI{0.002}{\degreeCelsius}), confirming low initialization
    sensitivity (see Section~\ref{sec:limitations} for details).}
  \label{fig:performance}
\end{figure*}

We conducted 30 holdout experiments across the four reefs, varying the holdout depth, the number of training depths, and the reef identity (Fig.~\ref{fig:performance}).
PINN holdout RMSE ranges from $0.25 \pm 0.001$\,\si{\degreeCelsius} (Davies Reef, \SI{9.1}{\metre} holdout) to $1.38 \pm 0.002$\,\si{\degreeCelsius} (Myrmidon Reef, \SI{7.3}{\metre} holdout with 3 training depths), where uncertainties are standard deviations across 5 random seeds.
Across all experiments, the PINN achieves the lowest RMSE in 8 of 30 cases (27\%).

This overall win rate understates the PINN's value because the wins concentrate in the scenarios that matter most for extending monitoring coverage.
At Myrmidon Reef, where the deep water column and offshore setting create strong vertical thermal gradients, the PINN wins 6 of 9 experiments (67\%), with RMSE reductions of up to 48\% relative to the best statistical baseline.
At the deepest Myrmidon holdout (\SI{14.7}{\metre}, full depth), the PINN achieves \SI{0.98}{\degreeCelsius} RMSE where the best statistical baseline (IDW) reaches \SI{1.35}{\degreeCelsius}, a \SI{0.37}{\degreeCelsius} improvement.
At Rib Reef's shallowest holdout (\SI{1.0}{\metre}, extrapolation from 6 and \SI{9}{\metre}), the PINN reaches \SI{0.41}{\degreeCelsius} against the best statistical baseline's \SI{0.84}{\degreeCelsius}, halving the error (though the FD physics-only baseline achieves \SI{0.39}{\degreeCelsius} at this configuration).

For shallow holdout depths where training data exist on both sides of the holdout, baselines are competitive or superior.
At Davies Reef with 17 training depths, RF achieves \SI{0.14}{\degreeCelsius} RMSE at the \SI{18.5}{\metre} holdout (where the adjacent \SI{18.1}{\metre} logger provides nearly co-located training data), versus the PINN's \SI{0.26}{\degreeCelsius}.
In these data-rich, interpolation-dominated scenarios, the physics constraint adds overhead without commensurate benefit; simple methods exploit nearby observations more efficiently.
The PINN's advantage emerges precisely when nearby observations are absent: at depth extremes, across gaps in the monitoring array, and under sparse coverage.

\subsection{Sparsity robustness}
\label{sec:sparsity}

\begin{figure}[!htbp]
  \centering
  \includegraphics[width=0.7\columnwidth]{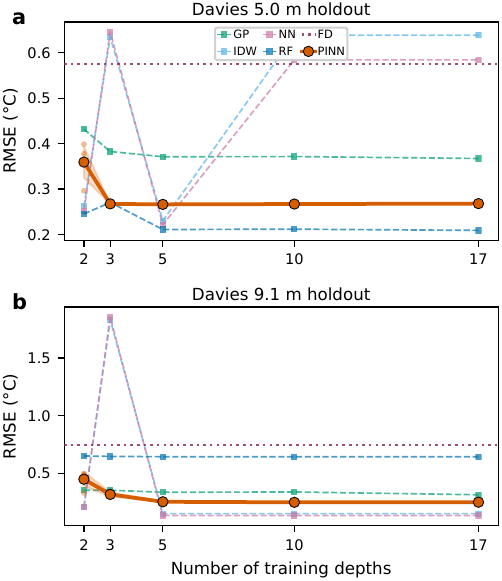}
  \caption{Reconstruction accuracy as a function of training data density for
    two Davies Reef holdout depths. The dotted horizontal line shows the
    physics-only FD baseline, which is constant (independent of training depth
    count). (a)~\SI{5.0}{\metre} holdout: PINN RMSE remains at
    $\sim$\SI{0.27}{\degreeCelsius} from 3 to 17 training depths, well below
    the FD baseline (\SI{0.58}{\degreeCelsius}), while statistical baselines
    degrade. (b)~\SI{9.1}{\metre} holdout: NN and IDW collapse from
    $\sim$\SI{0.13}{\degreeCelsius} to $>$\SI{1.8}{\degreeCelsius} at 3
    training depths, while the PINN holds at \SI{0.32}{\degreeCelsius}, again
    below the FD line (\SI{0.75}{\degreeCelsius}). The heat equation provides
    regularization that prevents catastrophic degradation under sparse data,
    and the neural network adds value beyond the physics alone. PINN lines
    show the mean across 5 seeds; translucent dots show individual seed RMSE
    values, and shaded bands show $\pm 1$ standard deviation.}
  \label{fig:sparsity}
\end{figure}

The PINN's most distinctive property is its stability under data sparsity (Fig.~\ref{fig:sparsity}).
At the Davies Reef \SI{5.0}{\metre} holdout, PINN RMSE remains between 0.266 and \SI{0.268}{\degreeCelsius} as training depths decrease from 17 to 3, rising to \SI{0.36}{\degreeCelsius} at 2 training depths.
Over the same range, RF degrades from 0.21 to \SI{0.27}{\degreeCelsius} and GP from 0.37 to \SI{0.38}{\degreeCelsius} (reaching \SI{0.43}{\degreeCelsius} at 2 training depths).

The pattern is more dramatic at the \SI{9.1}{\metre} holdout.
With 17 training depths, NN achieves \SI{0.13}{\degreeCelsius} (exploiting the adjacent \SI{8.5}{\metre} logger) and the PINN reaches \SI{0.25}{\degreeCelsius}.
But at 3 training depths, NN and IDW collapse to \SI{1.86}{\degreeCelsius} and \SI{1.83}{\degreeCelsius} respectively: the nearest training depth is now far from \SI{9.1}{\metre}, and without physics to guide interpolation, these methods fail catastrophically.
The PINN, by contrast, increases only to \SI{0.32}{\degreeCelsius}, a factor of 6 better than the collapsed baselines.
For reef managers, this means the PINN can reconstruct depth-resolved temperatures from as few as three logger depths with accuracy comparable to what baselines achieve only with dense arrays.

The physical explanation is straightforward.
The heat equation (Eq.~\ref{eq:heat}) constrains the vertical temperature profile to be consistent with diffusive mixing and depth-attenuated heating.
Even with just a surface and bottom observation, the physics determines the general shape of the vertical profile; the network fine-tunes within that physically constrained envelope.
Baselines lacking this constraint must infer the vertical structure entirely from data, which becomes impossible when observations are sparse.

\subsection{Depth-resolved thermal fields}
\label{sec:thermal_fields}

\begin{figure*}[!htbp]
  \centering
  \includegraphics[width=\textwidth]{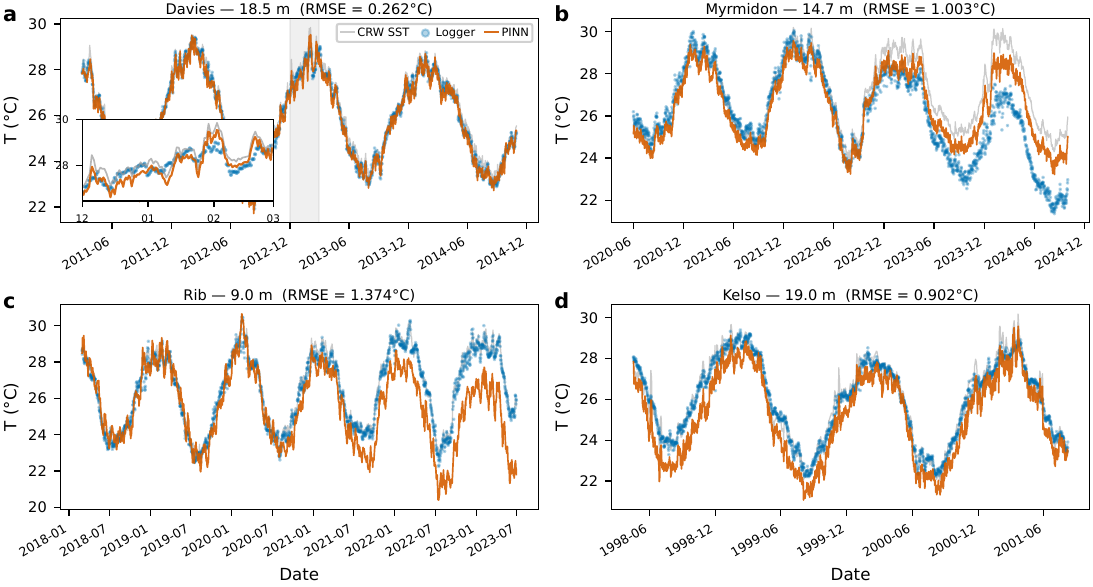}
  \caption{Time series comparison at four holdout depths: (a)~Davies Reef
    \SI{18.5}{\metre}, (b)~Myrmidon Reef \SI{14.7}{\metre}, (c)~Rib Reef
    \SI{9.0}{\metre}, (d)~Kelso Reef \SI{19.0}{\metre}. Gray: CRW satellite
    SST; blue dots: held-out logger observations; orange: PINN predictions.
    The inset in~(a) zooms into a December--February summer window where the
    PINN separates from SST and tracks the sub-surface logger.
    At Davies (17 training depths), the PINN captures the seasonal depth offset.
    At Myrmidon, Rib, and Kelso (5, 2, and 2 training depths
    respectively), the PINN remains pulled toward SST, particularly during
    summer peaks when the surface-to-depth offset is largest.}
  \label{fig:timeseries}
\end{figure*}

Figure~\ref{fig:timeseries} shows PINN-reconstructed temperature time series at four holdout depths alongside satellite SST and held-out logger observations.
The figure reveals both the method's strengths and a systematic limitation: the hard boundary condition (Eq.~\ref{eq:hard_bc}) anchors the PINN to satellite SST at the surface, and this anchor pulls the reconstruction toward SST at depth when training data are sparse.

At Davies Reef (\SI{18.5}{\metre}), the PINN tracks the logger closely (\SI{0.26}{\degreeCelsius} RMSE), reproducing both the summer maxima and the winter cooling that is stronger at depth than at the surface.
With 17 training depths spanning the full water column, the PINN has enough information to separate from the SST and reconstruct the true subsurface signal.
The satellite SST (gray line) overestimates temperature throughout the record, particularly during summer peaks.

The remaining three panels tell a different story.
At Myrmidon Reef (\SI{14.7}{\metre}), the PINN partially captures the surface-to-depth offset but remains pulled toward SST, particularly during the 2021--2023 summer peaks where the logger records temperatures 1--\SI{2}{\degreeCelsius} below SST while the PINN stays within \SI{0.5}{\degreeCelsius} of the satellite.
The \SI{1.00}{\degreeCelsius} RMSE reflects this incomplete separation.
At Rib Reef (\SI{9.0}{\metre}), the PINN is trained on only two depths (1 and \SI{6}{\metre}) and must extrapolate downward; the reconstruction tracks SST more closely than the logger, yielding \SI{1.37}{\degreeCelsius} RMSE.
At Kelso Reef (\SI{19.0}{\metre}), similarly trained on two depths (2 and \SI{7}{\metre}) far above the holdout, the PINN captures the seasonal cycle but underestimates the depth offset during summer, with \SI{0.90}{\degreeCelsius} RMSE.

The pattern across panels is consistent: the PINN separates from SST in proportion to the amount of subsurface training data available.
With dense depth coverage (Davies), the reconstruction is accurate; with sparse coverage and large extrapolation distances (Rib, Kelso), the hard BC dominates and the PINN defaults toward the satellite signal.
For reef managers, this means the PINN provides reliable depth correction at well-instrumented sites but should be interpreted cautiously when extrapolating far below the deepest training logger.

\subsection{Depth-resolved thermal stress}
\label{sec:thermal_stress}

\begin{figure*}[!htbp]
  \centering
  \includegraphics[width=\textwidth]{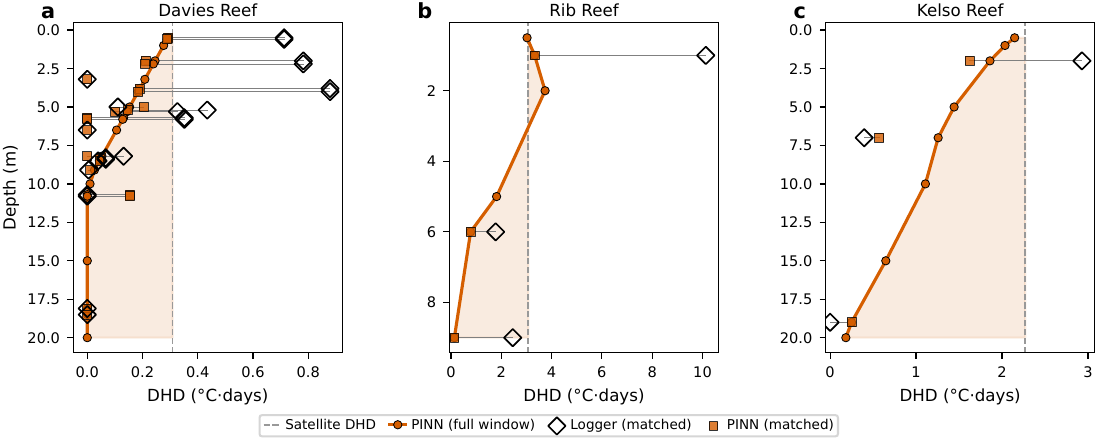}
  \caption{Depth-resolved thermal stress profiles. (a)~Davies Reef: cumulative
    Degree Heating Days (DHD, \si{\degreeCelsius\,days}) vs.\ depth. The dashed
    gray line shows satellite-derived DHD (constant with depth). The orange
    curve shows PINN-reconstructed DHD, which attenuates from the surface value
    with increasing depth. Diamond markers show logger-observed DHD computed
    over matched time windows. The shaded region shows the difference between
    satellite and PINN-reconstructed DHD at each depth. (b)~Rib Reef: the most
    dramatic depth attenuation, with DHD dropping from 3.33 at \SI{1}{\metre}
    to 0.14 at \SI{9}{\metre} while satellite DHD remains constant at 3.08.
    (c)~Kelso Reef: moderate thermal stress attenuating with depth.
    Myrmidon Reef is omitted because its maximum monthly mean (MMM) threshold
    of \SI{30.34}{\degreeCelsius} exceeded all observed temperatures during
    2020--2024, yielding DHD\,=\,0 at every depth; this is consistent with its
    exposed outer-shelf location where stronger currents and upwelling suppress
    thermal accumulation.}
  \label{fig:thermal_stress}
\end{figure*}

The central management question is whether thermal stress varies with depth, and if so, whether satellite SST captures that variation.
Figure~\ref{fig:thermal_stress} shows depth profiles of cumulative Degree Heating Days for Davies, Rib, and Kelso reefs, computed from logger observations, PINN predictions, and satellite SST over matched time windows.

Before examining the results, we note an architectural constraint that shapes the DHD comparison.
The hard boundary condition (Eq.~\ref{eq:hard_bc}) forces the PINN surface temperature to equal CRW SST at every timestep.
CRW SST is a daily nighttime-only foundation temperature product that excludes daytime observations to avoid solar contamination; it therefore does not capture daytime thermal peaks that in-situ loggers record at sub-hourly intervals.
The PINN therefore inherits the satellite's temporal smoothness at the surface and cannot produce DHD values exceeding the satellite-derived DHD near $z = 0$.
This constraint propagates downward: the PINN's depth-resolved DHD profile is bounded from above by the satellite value at the surface.

At Davies Reef (threshold = \SI{29.66}{\degreeCelsius}), this constraint is visible immediately.
The PINN surface DHD (0.29\,\si{\degreeCelsius\,days} at \SI{0.5}{\metre}) nearly equals the satellite DHD (0.31), while the logger at the same depth records 0.71\,\si{\degreeCelsius\,days}.
The logger captures sub-daily peaks that the daily satellite product (and therefore the PINN) cannot resolve.
At \SI{3.8}{\metre}, the pattern repeats: logger = 0.88, PINN = 0.19, satellite = 0.31.
The PINN underestimates stress not only because MSE training smooths predictions, but because the surface anchor itself lacks the temporal resolution to represent threshold exceedances.

At deeper depths ($>$\SI{8}{\metre}), however, the picture reverses.
Both PINN and loggers converge on zero DHD, confirming that thermal stress genuinely disappears at depth.
At the \SI{18.5}{\metre} logger (matched window: 1340 days), the logger records zero, the PINN predicts zero, and the satellite still reports 0.31\,\si{\degreeCelsius\,days}.
Here the PINN correctly removes the false positive stress that satellite SST assigns to deep water.

Rib Reef (threshold = \SI{30.05}{\degreeCelsius}) experienced active bleaching stress during its 2018--2023 study window and presents the starkest contrast.
The PINN reconstructs depth attenuation: DHD drops from 3.33 at \SI{1}{\metre} to 0.78 at \SI{6}{\metre} and 0.14 at \SI{9}{\metre}, while satellite DHD remains constant at 3.08\,\si{\degreeCelsius\,days}.
Logger observations, however, reveal higher stress than either method captures: 10.13\,\si{\degreeCelsius\,days} at \SI{1}{\metre}, 1.78 at \SI{6}{\metre}, and 2.46 at \SI{9}{\metre}.
At \SI{1}{\metre}, the logger DHD is 3$\times$ higher than both the PINN (3.33) and the satellite (3.08), indicating that high-frequency local thermal variability at this mid-shelf reef exceeds what the daily satellite grid can represent.
At \SI{9}{\metre}, the satellite (3.08) is closer to the logger (2.46) than the PINN is (0.14); the PINN overcompensates for the depth offset, underestimating stress by 94\%.

Kelso Reef shows intermediate behavior.
PINN DHD decreases from 2.15 at the surface to 0.25 at \SI{19}{\metre}, while satellite DHD persists at 2.27\,\si{\degreeCelsius\,days}.
At \SI{7}{\metre}, the logger records 0.40 and the PINN predicts 0.57, reasonable agreement.
At \SI{19}{\metre}, the logger records zero while the PINN predicts 0.25, a slight overestimation at depth.
At Myrmidon Reef, neither the PINN nor the satellite predict any thermal stress (DHD = 0 at all depths), because the MMM-based threshold (\SI{30.34}{\degreeCelsius}) exceeded observed temperatures throughout 2020--2024.
This absence of predicted stress during confirmed mass bleaching events (2020, 2024) likely reflects the spatial mismatch between the \SI{5}{\kilo\metre} CRW grid cell used for the MMM climatology and the reef's actual location, combined with Myrmidon's outer-shelf setting where temperatures may remain below the regional threshold even during bleaching events that affect the broader GBR.
Bleaching at Myrmidon may have been driven by the cumulative duration of moderate sub-threshold warming rather than by exceedance of MMM~+~\SI{1}{\degreeCelsius}; this warrants further investigation of threshold calibration at outer-shelf reefs.

Two conclusions emerge.
First, the PINN captures the qualitative pattern of thermal stress attenuation with depth, and at deep depths ($>$\SI{8}{\metre}) it correctly identifies zero stress where the satellite reports false positives.
Second, at shallow-to-mid depths, the PINN underestimates absolute DHD because its surface anchor (daily satellite SST) lacks the temporal resolution to capture the sub-daily thermal peaks that drive threshold exceedances.
PINN DHD profiles should be interpreted as conservative lower bounds on depth-resolved stress; the depth at which stress disappears is more reliable than the absolute DHD value at any given depth.

\subsection{Learned physical parameters}
\label{sec:parameters}

The PINN jointly estimates two physical parameters: effective thermal diffusivity ($\kappa$) and light attenuation coefficient ($K_d$).
Because these are learned rather than prescribed, they provide a consistency check: do the values fall within physically plausible ranges?

\begin{figure}[!htbp]
  \centering
  \includegraphics[width=\columnwidth]{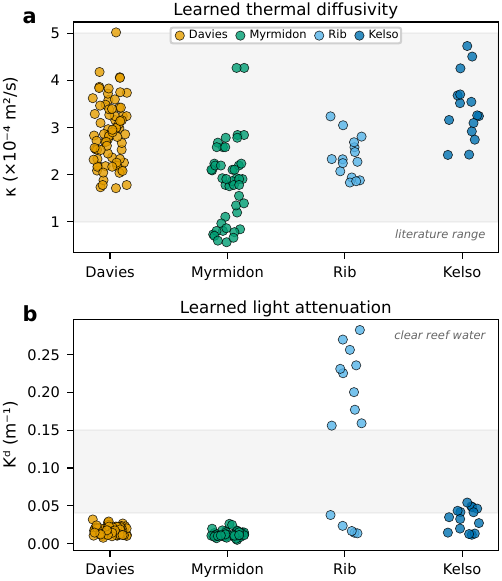}
  \caption{Learned physical parameters across reefs and holdout experiments
    (5 random seeds per configuration; each dot is one seed).
    (a)~Effective thermal diffusivity $\kappa$ ($\times 10^{-4}$\,\si{\metre
    \squared\per\second}). The gray band shows the literature range for
    turbulent mixing on coral reefs
    ($1$--$5 \times 10^{-4}$\,\si{\metre\squared\per\second};
    Monismith~\citep{monismith2007hydrodynamics}). Most learned values fall within or
    near this range. Tight clustering within configurations confirms low
    sensitivity to random initialization. (b)~Light attenuation coefficient $K_d$
    (\si{\per\metre}). The gray band shows the literature range for clear
    coral reef water (0.04--\SI{0.15}{\per\metre}; Kirk~\citep{kirk2011light}).
    Inter-reef variation in $K_d$ reflects differences in water clarity and
    depth coverage.}
  \label{fig:parameters}
\end{figure}

Figure~\ref{fig:parameters} shows the distribution of learned parameters across all holdout experiments and the full training-depth runs.
Effective diffusivities range from $\SI{0.6e-4}{}$ to $\SI{5.0e-4}{\metre\squared\per\second}$ across all experiments (150 PINN runs across 5 random seeds).
These values fall within the $10^{-4}$ to $10^{-3}$~\si{\metre\squared\per\second} range reported for turbulent mixing on coral reefs \citep{monismith2007hydrodynamics, lowe2015oceanic}.
Inter-reef variation is consistent with physical expectations: Myrmidon, an offshore reef with lower wave energy, yields the lowest full-data $\kappa$ ($\SI{1.27e-4}{\metre\squared\per\second}$), while Davies and Kelso, with greater exposure to tidal and wave-driven mixing, show higher full-data values ($\SI{4.37e-4}{}$ and $\SI{4.42e-4}{\metre\squared\per\second}$, respectively).

Light attenuation coefficients span a wider range.
Davies ($K_d = 0.008$--\SI{0.032}{\per\metre}) and Myrmidon ($K_d = 0.005$--\SI{0.026}{\per\metre}) show values mostly below the typical range for clear reef waters (0.04--\SI{0.15}{\per\metre}; Kirk~\citep{kirk2011light}), suggesting that the PINN may attribute some of the thermal attenuation to light attenuation when the two effects are difficult to separate from temperature data alone.
Kelso spans a wider range ($K_d = 0.012$--\SI{0.054}{\per\metre}), with most values near the lower edge of the literature band.
Rib Reef exhibits the highest variability ($K_d = 0.014$--\SI{0.283}{\per\metre}), likely reflecting the greater turbidity of this mid-shelf setting and the sensitivity of $K_d$ to the limited depth coverage (three depths).

\section{Discussion}
\label{sec:discussion}

\subsection{Implications for depth-resolved bleaching assessment}
\label{sec:bleaching_implications}

The central finding is that thermal stress varies with depth (by up to 75\% between 1 and \SI{9}{\metre} at Rib Reef), but satellite SST cannot capture this variation.
Both the PINN and logger observations confirm that cumulative thermal stress attenuates with depth. At Davies Reef, both PINN and loggers agree on zero DHD below \SI{10.7}{\metre} where the satellite still reports positive stress.
At Rib Reef, loggers record 10.1\,\si{\degreeCelsius\,days} at \SI{1}{\metre} but only 2.5 at \SI{9}{\metre}, a 75\% reduction that satellite SST (constant at 3.08) cannot resolve.

The PINN correctly captures the qualitative structure of this depth attenuation but underestimates absolute DHD values at shallow-to-mid depths, as discussed in Section~\ref{sec:thermal_stress}.
This underestimation arises because the PINN's smooth temperature reconstruction attenuates the short-duration thermal peaks that drive DHD accumulation, a consequence of optimizing for mean squared error rather than threshold exceedance metrics.
Future work could address this through peak-preserving loss functions or quantile regression.

Despite this limitation, three management implications emerge.
First, satellite-only bleaching alerts calibrated from SST \citep{liu2006overview, skirving2019relentless} apply the same stress estimate to all depths; both loggers and the PINN show this is incorrect, though the magnitude of the overestimation at depth remains uncertain.
Second, quantitative assessments of depth refugia \citep{bongaerts2010assessing, baird2018decline} require depth-resolved thermal profiles; even conservative (PINN-derived) profiles provide more information than the depth-uniform satellite assumption.
Third, the PINN's continuous depth representation enables direct comparison with post-bleaching surveys that record severity by depth \citep{bridge2013depth}, supporting more nuanced attribution of bleaching patterns.

The framework operates on existing infrastructure: satellite SST products and in-situ logger networks are already operational.
No new observational hardware is required; the method adds a physics-constrained computational layer that converts sparse point measurements into continuous depth-resolved fields.
An operational deployment would proceed as follows: for each monitored reef, a PINN is trained once on historical logger data and CRW SST (approximately \SI{4}{\minute} on a consumer GPU); the trained model then produces depth-resolved temperature and DHD predictions as new satellite SST data arrive, with no additional training required for inference.
Periodic retraining (e.g., seasonally) could incorporate new logger data.
For the ${\sim}$3,000 reefs in the GBR alone, training all PINNs in parallel would require approximately \SI{200}{GPU\text{-}hours}, a modest cost on modern cloud infrastructure.
The practical bottleneck is not computation but data availability: the framework requires at least 2--3 logger depths per reef, and most reefs globally lack any subsurface instrumentation.
Where logger networks similar to the AIMS program exist on other reef systems, the same framework could be applied with reef-specific training.

\subsection{When does the PINN outperform baselines?}
\label{sec:when_pinn_wins}

The PINN is not universally superior to statistical baselines; it wins 8 of 30 holdout experiments (27\%).
Understanding when and why the PINN provides added value is essential for practical deployment.

Three conditions favor the PINN.
First, deep holdout depths where no nearby training data exist benefit from the heat equation's constraint on vertical profile shape.
At Myrmidon Reef (\SI{14.7}{\metre} holdout, \SI{5}{} training depths), the PINN achieves \SI{0.98}{\degreeCelsius} RMSE versus the best statistical baseline's \SI{1.35}{\degreeCelsius}; the physics fills the gap between the nearest training depths (8 and \SI{20}{\metre}).
Second, sparse data regimes expose the baselines' reliance on local data density.
At the Davies \SI{9.1}{\metre} holdout with 3 training depths, baselines collapse to $>$\SI{1.8}{\degreeCelsius} RMSE while the PINN holds at \SI{0.32}{\degreeCelsius}.
Third, reefs with strong vertical gradients (Myrmidon, with its deep offshore water column) present a reconstruction challenge where physics-guided solutions are more reliable than statistical extrapolation.

Conversely, when training data are dense and the holdout depth is sandwiched between nearby loggers, statistical methods exploit proximity more efficiently than the PINN's physics.
RF and NN achieve \SI{0.14}{\degreeCelsius} RMSE at Davies \SI{18.5}{\metre} with the adjacent \SI{18.1}{\metre} logger in the training set; the PINN's \SI{0.26}{\degreeCelsius} RMSE reflects the overhead of satisfying the PDE constraint across the entire depth range.

These patterns suggest a practical decision framework.
The PINN is most valuable when (i)~the holdout depth lies beyond the convex hull of training depths (extrapolation), (ii)~training depth coverage is sparse ($\leq$5 loggers), or (iii)~the reef has a deep, stratified water column with strong vertical gradients.
When dense logger arrays already bracket the depth of interest, simpler interpolation methods (NN, IDW, RF) are computationally cheaper and equally or more accurate.
In practice, the PINN and baselines are complementary: a monitoring system could default to efficient statistical interpolation at well-instrumented depths and invoke the PINN for extrapolation to unmonitored depths.

More broadly, the PINN's value proposition scales with the ratio of physical knowledge to data availability.
Just as atmospheric correction of satellite imagery improves with physical models of radiative transfer \citep{donlon2012global}, depth correction of satellite SST improves with physical models of heat transfer.
The sparser the in-situ observations, the greater the relative benefit of the physics constraint.

\subsection{Limitations, uncertainty, and transferability}
\label{sec:limitations}

Several limitations qualify the results.

The one-dimensional formulation neglects lateral advection, internal waves, and tidal pumping.
On well-mixed reef platforms, vertical mixing dominates the thermal structure \citep{monismith2007hydrodynamics}, but reefs with strong lateral currents, tidal flushing, or internal wave activity may require a more complete physical model.
The absence of an explicit surface cooling term (longwave radiation, latent heat flux) means that the effective $\kappa$ absorbs cooling effects that are not purely diffusive; the learned values should be interpreted as effective parameters of the simplified model, not as direct measurements of turbulent diffusivity.

The matched-window DHD comparisons ensure temporal fairness, but logger deployment durations vary widely (29 to 1340 days across depths).
Short deployments yield DHD estimates that are sensitive to whether the deployment happened to coincide with a warm period.
We mitigate this by comparing all three sources (logger, PINN, satellite) over the same window, but the underlying temporal coverage heterogeneity should be considered when interpreting depth profiles.

Transferability to other reef systems requires validation.
Our four sites all lie in the central GBR, a region with relatively clear water and moderate wave exposure.
Reefs in the Caribbean, Southeast Asia, or the eastern Pacific may have different turbidity, mixing regimes, and depth ranges.
The framework itself is transferable: the heat equation is universal, and the PINN can be retrained with local data.
However, the learned parameters ($\kappa$, $K_d$) are reef-specific: across our four sites, $\kappa$ varies by an order of magnitude ($0.6$--$5.0 \times 10^{-4}$~\si{\metre\squared\per\second}) and $K_d$ by over an order of magnitude ($0.005$--\SI{0.28}{\per\metre}), reflecting real differences in mixing regimes and water clarity.
This variability suggests that transferring learned parameters directly between reefs without retraining would degrade accuracy substantially.
A more promising approach would be transfer learning: initializing a new reef's PINN with parameters learned at a similar reef (e.g., matching offshore vs.\ inshore setting), then fine-tuning with even a small amount of local logger data.
Whether this reduces the minimum data requirements below the 2--3 training depths tested here is an open question that warrants dedicated cross-reef experiments.

The DHD underestimation documented in Section~\ref{sec:thermal_stress} represents a fundamental tension between regression accuracy and threshold-based metrics.
The PINN is trained to minimize mean squared error on temperature, not to preserve extreme-value statistics.
Comparing PINN-derived DHD against matched-window logger observations reveals underestimation of 56--67\% at Rib Reef (1--\SI{6}{\metre}), 59--79\% at shallow Davies depths, and near-total suppression at Rib \SI{9}{\metre} (94\%) where the PINN's smooth reconstruction eliminates the short-duration thermal exceedances that drive cumulative stress.
A \SI{0.27}{\degreeCelsius} RMSE can yield large DHD bias when temperatures hover near the bleaching threshold, because even slight smoothing of peak temperatures can eliminate threshold crossings entirely.
Three approaches could address this limitation.
First, asymmetric or quantile loss functions that penalize underestimation of above-threshold temperatures more heavily than overestimation would directly target DHD accuracy.
Second, training on sub-hourly rather than hourly data would preserve more of the high-frequency thermal variability that drives threshold exceedances.
Third, ensemble approaches that sample from the PINN's posterior (e.g., via dropout or multi-seed aggregation) could provide exceedance probabilities rather than point estimates, enabling probabilistic DHD bounds.

To quantify initialization sensitivity, we repeated all 30 holdout experiments with 5 random seeds each (seeds 42--46), yielding 150 PINN training runs.
PINN RMSE is remarkably stable across seeds: the median coefficient of variation (CV) across all experiments is $\sim$0.6\%, with most configurations showing CV~$<$~2\%.
Higher variability occurs only under extreme sparsity: at Davies \SI{9.1}{\metre} with 2 training depths, CV reaches 13.6\% (RMSE $0.447 \pm 0.061$\,\si{\degreeCelsius}), and at Kelso \SI{19.0}{\metre} with 2 training depths, CV reaches 12.8\% (RMSE $0.753 \pm 0.096$\,\si{\degreeCelsius}).
In well-instrumented configurations ($\geq$5 training depths), CV is consistently below 1.5\%.
These results confirm that PINN performance differences reported in this study reflect genuine method capabilities rather than initialization artifacts.

We also compared the PINN against a physics-only finite-difference (FD) baseline: the same 1D heat equation solved via implicit Euler with literature-value parameters ($\kappa = \SI{2.5e-4}{\metre\squared\per\second}$, $K_d = \SI{0.1}{\per\metre}$) and real CRW SST as the surface boundary condition, but no neural network and no data assimilation beyond the surface.
The PINN outperforms the FD baseline in 27 of 30 experiments (90\%).
The FD baseline produces 2--4$\times$ higher RMSE than the PINN at most configurations (e.g., Davies \SI{18.5}{\metre}: FD = \SI{1.06}{\degreeCelsius} vs.\ PINN = \SI{0.26}{\degreeCelsius}; Myrmidon \SI{14.7}{\metre}: FD = \SI{2.07}{\degreeCelsius} vs.\ PINN = \SI{0.98}{\degreeCelsius}).
The FD baseline wins only at Rib Reef's shallowest and deepest holdouts (1.0 and \SI{9.0}{\metre}) and Kelso's \SI{2.0}{\metre} holdout, all cases where the fixed-parameter physics happens to match local conditions well.
This confirms that the neural network component adds clear value: it adapts the effective parameters to local conditions and assimilates subsurface observations, capabilities that a pure physics model with literature parameters cannot match.

No ablation study was conducted for hyperparameters such as the number of Fourier features, the PDE weight schedule, or the network architecture.
The sensitivity of results to these choices remains to be quantified.

Training on sub-hourly rather than hourly data could also improve DHD accuracy by preserving more high-frequency thermal variability, at the cost of increased training time and memory requirements.

\subsection{Future directions}
\label{sec:future}

Five extensions are natural.
First, the DHD-aware loss functions and ensemble approaches outlined above could make the PINN suitable for quantitative bleaching risk assessment rather than only qualitative depth profiling.
Second, incorporating additional remote sensing products (satellite-derived chlorophyll-$a$ as a prior for $K_d$, wind speed as a proxy for near-surface mixing) would constrain the learned parameters and may improve extrapolation to unmonitored depths.
Third, extension to two dimensions (depth plus cross-reef horizontal distance) would capture lateral temperature gradients on reef slopes.
Fourth, systematic cross-reef transfer experiments (training on a well-instrumented reef and fine-tuning with minimal local data at a new site) would quantify how few loggers are truly needed for deployment and whether learned $\kappa$ and $K_d$ values provide useful initializations.
Fifth, coupling depth-resolved thermal fields with coral bleaching probability models would produce spatially and vertically explicit bleaching risk maps, moving beyond the binary satellite alert paradigm.

\section{Conclusions}
\label{sec:conclusions}

Satellite sea surface temperature is the foundation of global coral bleaching monitoring, but it sees only the surface.
We have shown that a physics-informed neural network fusing CRW satellite SST with sparse AIMS temperature loggers can reconstruct depth-resolved thermal fields across four Great Barrier Reef sites with holdout accuracy of 0.25--\SI{1.38}{\degreeCelsius} RMSE.
The PINN maintains stable accuracy under extreme data sparsity: at three training depths, it achieves \SI{0.27}{\degreeCelsius} RMSE at the \SI{5}{\metre} holdout and \SI{0.32}{\degreeCelsius} at the \SI{9.1}{\metre} holdout, where statistical baselines collapse to $>$\SI{1.8}{\degreeCelsius}.
Multi-seed experiments (5 seeds per configuration) confirm low initialization sensitivity (median CV~$\sim$0.6\%), and the PINN outperforms a physics-only finite-difference baseline in 90\% of experiments, demonstrating that the neural network provides value beyond the governing equations with fixed parameters.
The learned effective thermal diffusivities ($0.6$--$5.0 \times 10^{-4}$~\si{\metre\squared\per\second}) fall within the literature range for turbulent mixing on coral reefs, confirming the physical consistency of the reconstruction.

Depth-resolved DHD profiles reveal that thermal stress attenuates with depth. This pattern is confirmed by both PINN predictions and in-situ logger observations but remains invisible to satellite SST.
At Davies Reef, both PINN and loggers agree on zero thermal stress below \SI{10.7}{\metre} where the satellite reports positive stress at all depths.
At Rib Reef, loggers record a 75\% reduction in DHD between \SI{1}{\metre} and \SI{9}{\metre}, while satellite SST remains constant.
However, matched-window validation reveals that the PINN underestimates absolute DHD relative to loggers at shallow-to-mid depths, because the smooth PINN reconstruction attenuates short-duration temperature peaks that drive threshold exceedances.
The PINN should therefore be interpreted as providing the \emph{shape} of the depth-stress profile (a conservative lower bound) rather than unbiased DHD estimates.
For reef managers, the key result is that applying satellite SST uniformly to all depths misrepresents subsurface thermal exposure, with implications for depth refugia assessment, alert threshold calibration, and the interpretation of depth-stratified bleaching surveys.

The framework requires no new observational infrastructure.
Satellite SST products and in-situ logger networks already exist at numerous reef systems globally.
Physics-constrained data fusion provides a pathway to extend these observations to the depth dimension, delivering the depth-resolved thermal information needed to understand bleaching patterns in three dimensions as marine heatwaves intensify under climate change.

\section*{Data and code availability}

Temperature logger data are available from the AIMS Data Centre (\url{https://data.aims.gov.au}).
CRW satellite SST data are available from NOAA CoastWatch ERDDAP (\url{https://coastwatch.pfeg.noaa.gov/erddap/}).
Code for the PINN framework and all experiments will be made available upon publication.

\section*{CO$_2$ emissions related to experiments}

All experiments were conducted on a single NVIDIA RTX~4090 GPU (TDP \SI{450}{\watt}).
The 150 multi-seed holdout validation runs (30 configurations $\times$ 5 seeds) required \SI{26658}{\second} (\SI{7.4}{\hour}) of GPU time, and the 4 thermal stress runs required \SI{973}{\second} (\SI{0.27}{\hour}), for a total of \SI{7.7}{GPU\text{-}hours}.
Including an estimated \SI{100}{\watt} for CPU and memory overhead, total energy consumption was approximately \SI{4.2}{\kilo\watt\hour}.
Using the Australian average grid emission factor of \SI{0.68}{\kilogram\,CO_2\per\kilo\watt\hour}, we estimate total emissions of approximately \SI{2.9}{\kilogram} CO$_2$-equivalent for all reported experiments.
This is comparable to driving an average car for \SI{12}{\kilo\metre}.
Preliminary development runs (hyperparameter tuning, architecture selection, debugging) are estimated to have consumed an additional \SIrange{5}{10}{GPU\text{-}hours}, bringing total project emissions to approximately \SIrange{5}{8}{\kilogram} CO$_2$-equivalent.

\section*{Author Contributions}

Alzayat Saleh: Conceptualization, methodology, software, data curation, formal analysis, investigation, writing (original draft), visualization, and project administration.
Mostafa Rahimi Azghadi: Conceptualization and review \& editing of the manuscript.

\section*{Declaration of competing interest}

The authors declare that they have no known competing financial interests or personal relationships that could have appeared to influence the work reported in this paper.

\section*{Funding}

This research did not receive any specific grant from funding agencies in the public, commercial, or not-for-profit sectors.

\section*{Acknowledgements}

The authors thank collaborators at James Cook University, the Australian Institute of Marine Science (AIMS), and the AIMS@JCU partnership.
Computational resources were provided by James Cook University's High Performance Computing facilities.
The authors used Generative AI to assist with manuscript drafting.
All experimental results, scientific interpretations, and final text were reviewed and verified by the authors, who take full responsibility for the content.

\bibliographystyle{elsarticle-num}
\bibliography{references}

\end{document}